%% file: main.tex
\documentclass{isprs}
\usepackage{natbib}
\usepackage{subfigure}
\usepackage{setspace}
\usepackage{geometry}
\usepackage{xargs}
\usepackage[pdftex,dvipsnames]{xcolor}
\usepackage{blindtext}
\usepackage[inline]{enumitem}
\usepackage{xcolor}
\usepackage{hyperref}
\usepackage{etoolbox}
\usepackage{algorithm}
\usepackage{algpseudocode}
\usepackage{amsmath}
\usepackage{placeins}
\usepackage{float}
\usepackage{flushend} 
\usepackage{enumitem}

\newcommandx{\simpletodo}[2][1=]{\todo[inline,linecolor=Black,#1]{Simpletodo: #2}}
\newcommandx{\todoUnsure}[2][1=]{\todo[linecolor=red,backgroundcolor=red!25,bordercolor=red,#1]{Unsure: #2}}
\newcommandx{\todoChange}[2][1=]{\todo[linecolor=blue,backgroundcolor=blue!25,bordercolor=blue,#1]{Change: #2}}
\newcommandx{\todoInfo}[2][1=]{\todo[linecolor=OliveGreen,backgroundcolor=OliveGreen!25,bordercolor=OliveGreen,#1]{Info: #2}}
\newcommandx{\todoImprovement}[2][1=]{\todo[linecolor=Plum,backgroundcolor=Plum!25,bordercolor=Plum,#1]{Improvement: #2}}

\newcommandx{\todoTable}[2][1=]{\todo[inline,linecolor=red,backgroundcolor=red!25,bordercolor=red,#1]{Table: #2}}
\newcommandx{\todoPlot}[2][1=]{\todo[inline,linecolor=OliveGreen,backgroundcolor=OliveGreen!25,bordercolor=OliveGreen,#1]{Plot: #2}}

\begin{document}
\title{%
Deep Learning Enhanced Road Traffic Analysis: \\
Scalable Vehicle Detection and Velocity Estimation Using PlanetScope Imagery}

\author{
Adamiak Maciej\textsuperscript{1,3* },
Yulia Grinblat\textsuperscript{1* },
Julian Psotta\textsuperscript{1,2* },
Nir Fulman\textsuperscript{1,2 },
Himshikhar Mazumdar\textsuperscript{1 },
Shiyu Tang\textsuperscript{2 } and
Alexander Zipf\textsuperscript{1,2 }
}

\address
{
\textsuperscript{1 }Heidelberg Institute for Geoinformation Technology (HeiGIT) Heidelberg, Germany\\
\textsuperscript{2 }GIScience Research Group, Institute of Geography, Heidelberg University, Germany\\
\textsuperscript{3 }Institute of Urban Geography, Tourism Studies and Geoinformation; Faculty of Geographical Sciences; University of Lodz, Poland 
\\
\textsuperscript{* }Corresponding author
\\
\begin{itemize*}[font={\color{white}}]
	\item \color{blue} \href{mailto:maciej.adamiak@geo.uni.lodz.pl}{maciej.adamiak@geo.uni.lodz.pl}
	\item \color{blue} \href{mailto:yulia.grinblat@heigit.org}{yulia.grinblat@heigit.org}
	\item \color{blue} \href{mailto:julian.psotta@heigit.org}{julian.psotta@heigit.org}
\end{itemize*}
}

\abstract{This paper presents a method for detecting and estimating vehicle speeds using PlanetScope SuperDove satellite imagery, offering a scalable solution for global vehicle traffic monitoring. Conventional methods such as stationary sensors and mobile systems like UAVs are limited in coverage and constrained by high costs and legal restrictions. Satellite-based approaches provide broad spatial coverage but face challenges, including high costs, low frame rates, and difficulty detecting small vehicles in high-resolution imagery. We propose a Keypoint R-CNN model to track vehicle trajectories across RGB bands, leveraging band timing differences to estimate speed. Validation is performed using drone footage and GPS data covering highways in Germany and Poland. Our model achieved a Mean Average Precision of 0.53 and velocity estimation errors of approximately 3.4 m/s compared to GPS data. Results from drone comparison reveal underestimations, with average speeds of 112.85 km/h for satellite data versus 131.83 km/h from drone footage. While challenges remain with high-speed accuracy, this approach demonstrates the potential for scalable, daily traffic monitoring across vast areas, providing valuable insights into global traffic dynamics.}


\maketitle
\input{chapters/01Introduction}
\input{chapters/02RelatedWorks}
\input{chapters/03MaterialsAndMethods}

\input{chapters/04Results}

\input{chapters/05DiscussionAndConclusion}
\input{chapters/06AuthorContributions}
\input{chapters/07Acknowledgement}
\input{chapters/08DeclarationOfCompetingInterest}
\input{chapters/09References}
\twocolumn

\end{document}

%% file: chapters/01Introduction.tex
\section{Introduction}
\label{sec:01Introduction}
Accurate traffic speed measurement is vital for managing transportation systems, improving safety, efficiency, and reducing congestion. While conventional traffic data such as loop detectors, cameras, and radar are effective, they are costly and limited in scale, restricting their use for real-time applications like vehicle routing. These systems often rely on static speed limits and models, resulting in less reliable time estimates compared to real-time traffic data. Other traffic data collection methods from mobile sensors, like UAVs and GPS-equipped probe cars, offer wider coverage but face challenges, including privacy concerns, spatial inaccuracies, and limited data in dense urban areas.
\\
Satellite-based approaches offer extensive, continuous coverage for traffic monitoring, overcoming the limitations of conventional stationary and mobile sensors. Several works used high-resolution satellite video to detect vehicles and estimate their speeds. Still, this approach is costly, lacks scalability, and is prone to errors due to low frame rates and small foreground sizes. Other studies have used very high-resolution ($\sim$1m) images for vehicle detection, with early works relying on simple machine learning models, while more recent methods employ convolutional neural networks for more accurate detection. However, scaling based on very high-resolution images remains a challenge.
\\
High-resolution sensors like PlanetScope SuperDove ($\sim$3.7m resolution)  provide global and daily coverage \citep{planet2024planetscope}. Various works have developed vehicle detection methods that rely on highlighting them against the background \citep{drouyer2019highway, chen2021spatial}. Recent studies have utilized medium-resolution imagery to detect fast-moving objects, leveraging the band-shifting effect caused by the satellite’s movement and the push-broom sensor to estimate object velocities. While some studies have used manual methods to assess the speed of objects in satellite images, these methods are difficult to scale \citep{kaab2014motion, binet2022accurate, heiselberg2019aircraft, heiselberg2021aircraft, keto2023detection}. Recently, \citet{fisser2022detecting} presented an automatic approach using multispectral Sentinel-2 data and a random forest classifier to detect moving trucks. However, their method struggles with low spectral contrast between vehicles and the road, particularly on reflective surfaces or in visually complex environments.
\\
We suggest an alternative approach that overcomes these challenges by employing a Keypoint R-CNN model to detect and track vehicle trajectories across multiple spectral bands, focusing on spatial structure rather than spectral contrast alone. Our method identifies keypoints representing vehicle geometry, tracking them across the blue, red, and green bands, enabling consistent motion detection even when spectral differences are minimal. We utilize PlanetScope SuperDove imagery \citep{planet2024planetscope}, which offers daily global coverage at a resolution of approximately 3.7 metres, enabling us to detect individual vehicles with precision. Moreover, our approach incorporates a novel validation process using speed estimates from drone footage and GPS tracks, an improvement over previous methods that lacked such detailed comparison.

%% file: chapters/02RelatedWorks.tex
\section{Related works}
\label{sec:02RelatedWorks}
Traffic data is most commonly collected from stationary sensors. These include loop detectors, road traffic cameras, and more recently, radar and laser sensors. While loop detectors provide velocity data directly, extracting this information from surveillance cameras requires analyzing images or videos, which have become prolific due to advancements in deep learning \citep{zhang2017fcn}. Stationary tools enable continuous monitoring of traffic conditions. However, they are fixed to specific locations, and installing such detectors across an entire city is challenging due to their high installation and operational expenses. As a result, they typically do not cover large areas \citep{peppa2018urban}.
\\
Mobile traffic sensors, such as UAVs (Unmanned Aerial Vehicles) and GPS-equipped probe cars, have become valuable tools for traffic monitoring due to their mobility, adaptability and low cost. UAVs can detect and track moving vehicles through machine learning algorithms, leveraging techniques such as object detection, classification, and trajectory prediction. However, a single UAV can only cover a limited space, and is also limited by legal restrictions on usage, especially in built-up areas \citep{ke2018real, barmpounakis2020new}. Similarly, GPS data collected from probe cars or smartphones offers broad spatial coverage and accurate speed information. However, the penetration rate remains low because only a small percentage of vehicles use GPS devices or share location data, often due to privacy concerns, inconsistent app adoption, and gaps in coverage. While this data is generally reliable for measuring vehicle speed, it is insufficient for capturing crucial metrics like traffic volume and density \citep{herrera2010evaluation}. Additionally, GPS car-probe studies face challenges such as spatial inaccuracies due to inherent GPS limitations, especially in built-up areas, and discontinuous records across spatial and temporal scales \citep{seo2017traffic}.
\\
Satellite-based approaches promise to provide extensive and continuous spatial coverage in traffic flow monitoring, overcoming the limitations of conventional stationary and mobile sensors. One line of work focuses on satellite video analysis, where high-resolution imagery from commercial SkySat sensors \citep{kopsiaftis2015vehicle,pflugfelder2020} is used to monitor traffic. These studies employed machine learning methods to detect vehicles and estimate their speeds. However, this approach is costly, lacks scalability, and suffers from low frame rates and small foreground sizes, leading to frequent errors.\\
Some studies have utilized very high-resolution satellite images for vehicle detection. Early works, like \citet{hinz2005context}, used geometric line extraction to identify vehicle queues, including traffic congestion and parked cars, while \citet{larsen2009traffic} improved detection by adding vehicle shadows through image segmentation, followed by feature extraction and classification. Recently, the focus has shifted toward more advanced machine learning models and convolutional neural networks (CNNs). Two-stage models like R-CNNs \citep{xu2017car,zhang2017fcn,peppa2018urban,ke2018real} identify and refine potential vehicles, while one-stage detectors such as YOLO offer faster, real-time detection with higher accuracy \citep{bozcan2020au,sun2022rsod}. Despite their effectiveness, very high-resolution images present challenges when scaling up for larger areas. Other works have used medium-resolution sensors with global coverage for traffic counts. For instance, \citet{drouyer2019highway} and \citet{chen2021spatial} detected vehicles using a technique called multi-directional top-hat transformation. This method highlights bright objects, like vehicles, by subtracting the background from satellite images, allowing for easier detection of the residual images, such as those derived from PlanetScope imagery.
\\
Detecting moving objects in satellite images is possible by leveraging the time-lag between sensor bands in push broom systems \citep{kaab2014motion, binet2022accurate}. This approach works by using the slight delay between the acquisition of bands within a single satellite image to detect movement. Some studies have used manual methods to assess the speed of objects in satellite images. \citet{heiselberg2019aircraft} and \citet{heiselberg2021aircraft} exploited timing differences in MSI Sentinel-2 images to detect aircraft and ship motion by applying statistical thresholds to distinguish target pixels from the background. \citet{keto2023detection} used PlanetScope images to detect aircraft movement and estimate speed through band-specific image thresholding. However, manual methods are difficult to scale up.
\\
\citet{fisser2022detecting} proposed an automatic method for detecting moving trucks on roads using Sentinel-2 data, which exploits the temporal sensing offsets of the multispectral instrument to capture spatial and spectral distortions caused by object movement. Their approach uses a random forest classifier trained on spectral differences between vehicles and their background, achieving an overall accuracy of 84\%. While effective in certain conditions, the method faces challenges when the spectral contrast between vehicles and the road is minimal, such as on reflective surfaces or in complex environments where the background is visually similar. These limitations make it difficult to reliably distinguish vehicles from their surroundings, particularly when spectral differences alone are not pronounced. 
\\
We propose a Keypoint R-CNN model to detect vehicle trajectories across multiple spectral bands by identifying keypoints that represent the vehicles' spatial structure and tracking them across blue, red, and green bands. By linking these keypoints over time, we model the vehicle's geometric consistency and movement, even when spectral differences are minimal.

%% file: chapters/03MaterialsAndMethods.tex
\section{Materials and Methods}
\label{sec:03MaterialsAndMethods}
\input{chapters/03MaterialsAndMethods/03_01Datasets}
\input{chapters/03MaterialsAndMethods/03_02ConceptualFrameworkOnMovingEchoes}
\input{chapters/03MaterialsAndMethods/03_03TheProposedAlgorithm}
\input{chapters/03MaterialsAndMethods/03_04VelocityEstimation}
\input{chapters/03MaterialsAndMethods/03_05VelocityValidation}

%% file: chapters/03MaterialsAndMethods/03_01Datasets.tex
\subsection{Datasets}
\label{subsec:03Datasets}
\subsubsection{PlanetScope Imagery}
\label{subsubsec:03PlanetScopeImagery}
The PlanetScope satellite constellation is an Earth observation system operated by Planet Labs \citep{planet2024planetscope}. It consists of over 130 satellites capable of monitoring the Earth's entire land surface daily \citep{roy2021global}. The latest deployment of PlanetScope satellites, designated PSB.SD and referred to as SuperDove, enables capturing medium-resolution imagery ($\sim$3.7m/px Ground Sampling Distance (GSD)) with an imager frame consisting of 8880px x 5280px, 660px for each band respectively \citep{saunier2022technical}. The ordering of bands in the imager frame is blue (490 nm), red (665 nm), green 1 (531 nm), green (565 nm), yellow (610 nm), red edge (705 nm), near-infrared (865 nm), coastal blue (442 nm). The deployment allows for the acquisition of images daily between 9:00 and 11:00 am for the entire planet.
\\
We acquired Planet Scope 3-bands (RGB) satellite images that cover the following highways with a minimum of 50\% visibility and a maximum of 50\% cloud cover:
\begin{itemize}[noitemsep, topsep=0pt]
    \item[--] Germany A2: Hannover to Lehnin — Length: 215 km; Capture Date: 26.09.2023.
    \item[--] Germany A7: Hannover to Kassel — Length: 172 km; Capture Date: 26.09.2023.
    \item[--] Poland A1: Plichtow Segment — Length: 620 m; Location Coordinates: 51°49’N, 19°38’E; Capture Dates: between 16.03.2023 and 15.03.2024.
    \item[--] Germany: Images that match 58 GPS tracks (details below). Capture Dates: between 2020-03-15 and 2024-04-25.
\end{itemize}
\subsubsection{Drone footage}
\label{subsubsec:03DroneFootage}
\begin{figure*}
    \centering
    \includegraphics[width=1\linewidth]{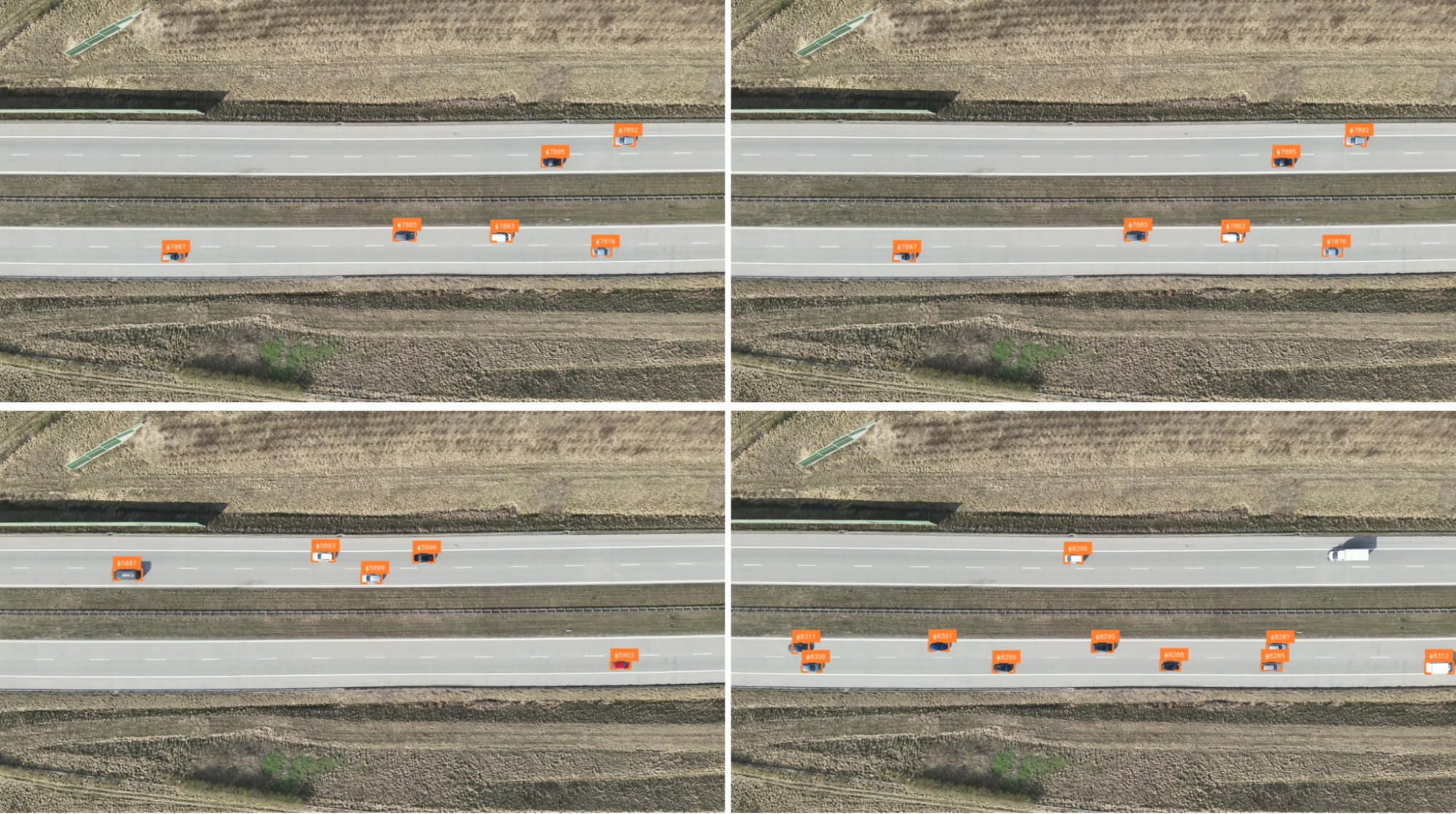}
    \caption{Drone footage vehicle tracking examples acquired on the A1 highway near Plichtow, Poland.}
    \label{fig:DroneFootageVehicleTracking}
\end{figure*}

Drone video footage is utilized to calculate and compare vehicle speed distributions along a road segment, for the purpose of validating the analysis conducted using satellite imagery. We employed drone video footage from a DJI Mavic 3T (M3T) at a frame rate of 30 fps to calculate the speed distribution across a selected 620m segment of the Polish A1 highway during the acquisition of SuperDove satellite imagery from 09:00 to 10:00 am. Over one hour of video material from March 16, 2024, was collected (see Figure \ref{fig:DroneFootageVehicleTracking}).
\subsubsection{GPS tracks}
\label{subsubsec:03GpsTracks}
GPS tracks are used to provide a comparable estimate of individual vehicle speeds, enabling validation of the velocities derived from satellite imagery. These trajectories are crowdsourced by volunteers in the enviroCar platform \citep{broring2015envirocar} and provided as open data \cite{envirocar_nd}. Speed estimates, calculated from the displacement between sequential GPS coordinates, are included with the dataset. The original data set comprises over twenty thousand individual trajectories recorded by over one thousand vehicles between 15 March 2020 to 30 April 2024. The time resolution varies, but on average, it is $\sim$5 sec, depending on the type of GPS sensor. The GPS data primarily consists of tracks recorded in Germany, California, and Florida states (USA), with a smaller number of trajectories collected in India.
\subsubsection{OpenStreetMap (OSM) road data}
\label{subsubsec:03OpenStreetMapRoadData}
We extracted road data to delineate the highways within our satellite imagery using OpenStreetMap (OSM) data. This data was obtained via Overture \citep{overture2024overture}, a platform that facilitates the integration of detailed global GIS layers.

%% file: chapters/03MaterialsAndMethods/03_02ConceptualFrameworkOnMovingEchoes.tex
\subsection{Conceptual framework on ‘moving echoes’}
\label{subsec:03ConceptualFrameworkOnMovingEchoes}
\begin{figure}[H]
    \centering
    \includegraphics[width=1\columnwidth]{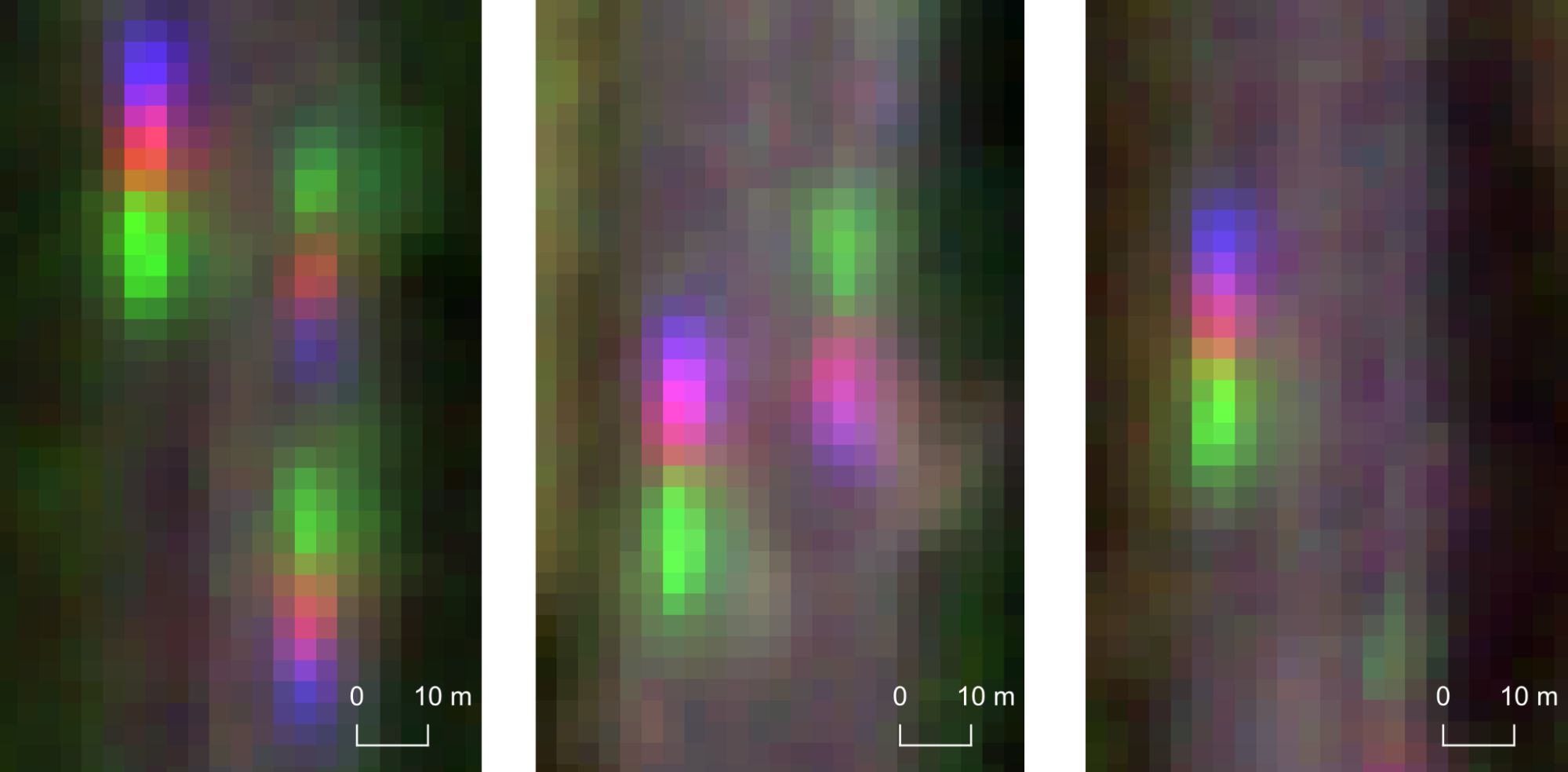}
    \caption{‘Moving echoes’ on true colour SuperDove satellite imagery on the A7 highway; Hanover-Kassel, Germany.}
    \label{fig:MovingEchoes}
\end{figure}

PlanetScope SuperDove \citep{planet2024planetscope} operates using a push broom scanner, meaning that the order of the band filters is also the order in which band images are recorded over time. Rapidly moving vehicles appear blurry in images in SuperDove images. The blurring is caused by the sensor recording each colour band at slightly different times, capturing the moving vehicle in different positions across the bands. Each moving vehicle in an image band appears as an elongated cluster of pixels that typically peak in intensity toward the centre, creating a visual effect of pixel shifts between the colour bands. We refer to this phenomenon as the ‘object in motion echo’. 
\\
This is illustrated in Figure \ref{fig:MovingEchoes}, which shows multiple vehicles on roadways, captured as blurry lines. Each vehicle is represented with a red, green, and blue colour gradient, showing the sequential captures across the satellite's sensor images. The blue end of each shape indicates the initial position, the red represents the mid-sequence, and the green marks the final position, illustrating the direction and extent of movement during the image acquisition process. 
\\
We measure vehicle velocities by analyzing the displacement across blue, red, and green band captures, exploiting the timing differences to detect positions and trajectories. We use only these bands because vehicles are most easily distinguished by the human eye in these wavelengths, making them easier to identify and annotate. 

%% file: chapters/03MaterialsAndMethods/03_03TheProposedAlgorithm.tex
\subsection{The proposed algorithm}
\label{subsec:03TheProposedAlgorithm}
\begin{figure*}[t]
    \centering
    \includegraphics[width=1\linewidth]{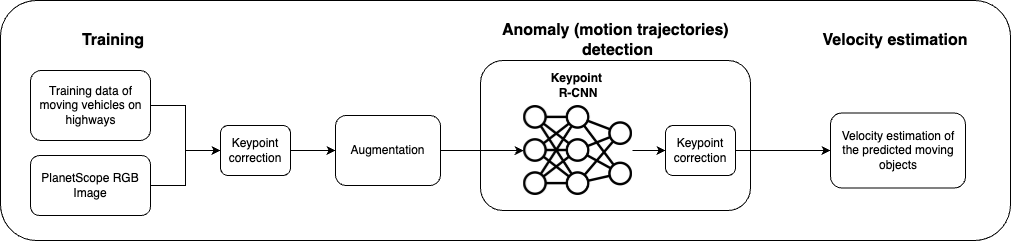}
    \caption{An overview of the proposed algorithm for detecting moving vehicles and estimating their speed based on SuperDove images.}
    \label{fig:ProposedAlgorithm}
\end{figure*}
Our approach to analyzing vehicle motion in PlanetScope imagery is structured into two steps. Initially, we detect the position and trajectory of vehicles by examining the positional shifts across the SuperDove blue, red, and green bands. This detection is framed as a pose estimation problem \citep{Zheng2023poseestimation}, where each vehicle is represented by three keypoints, sequentially linked across the bands, and analyzed using a Keypoint R-CNN model to predict their trajectories. Subsequently, we estimate the velocity of the vehicles by measuring the distance they travel along the keypoint-defined trajectories, relative to the time interval between the recorded bands. Figure \ref{fig:ProposedAlgorithm} presents a flowchart of the algorithm used to train for detecting and estimating the speed and direction of moving vehicles using SuperDove images. The subsequent sections detail the components of our algorithm and describe the validation of each step.

\subsubsection{Detecting vehicle trajectories}
\label{subsubsec:03DetectingVehicleTrajectories}
\begin{figure}[H]
    \centering
    \includegraphics[width=1\columnwidth]{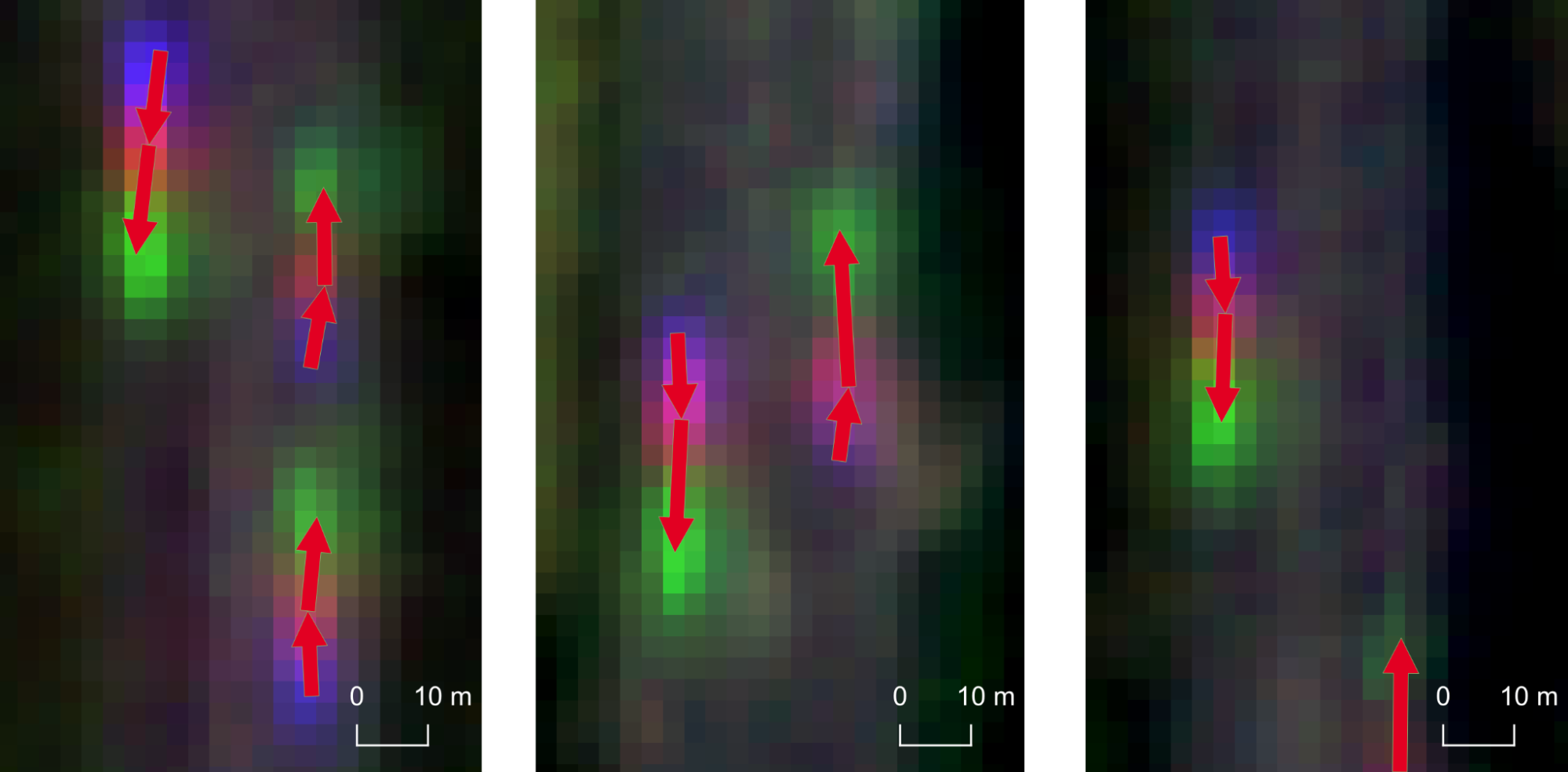}
    \caption{‘Moving echoes’ with trajectory training labels.}
    \label{fig:MovingEchoesWithTrajectories}
\end{figure}

\begin{figure*}[t]
    \centering
    \includegraphics[width=1\linewidth]{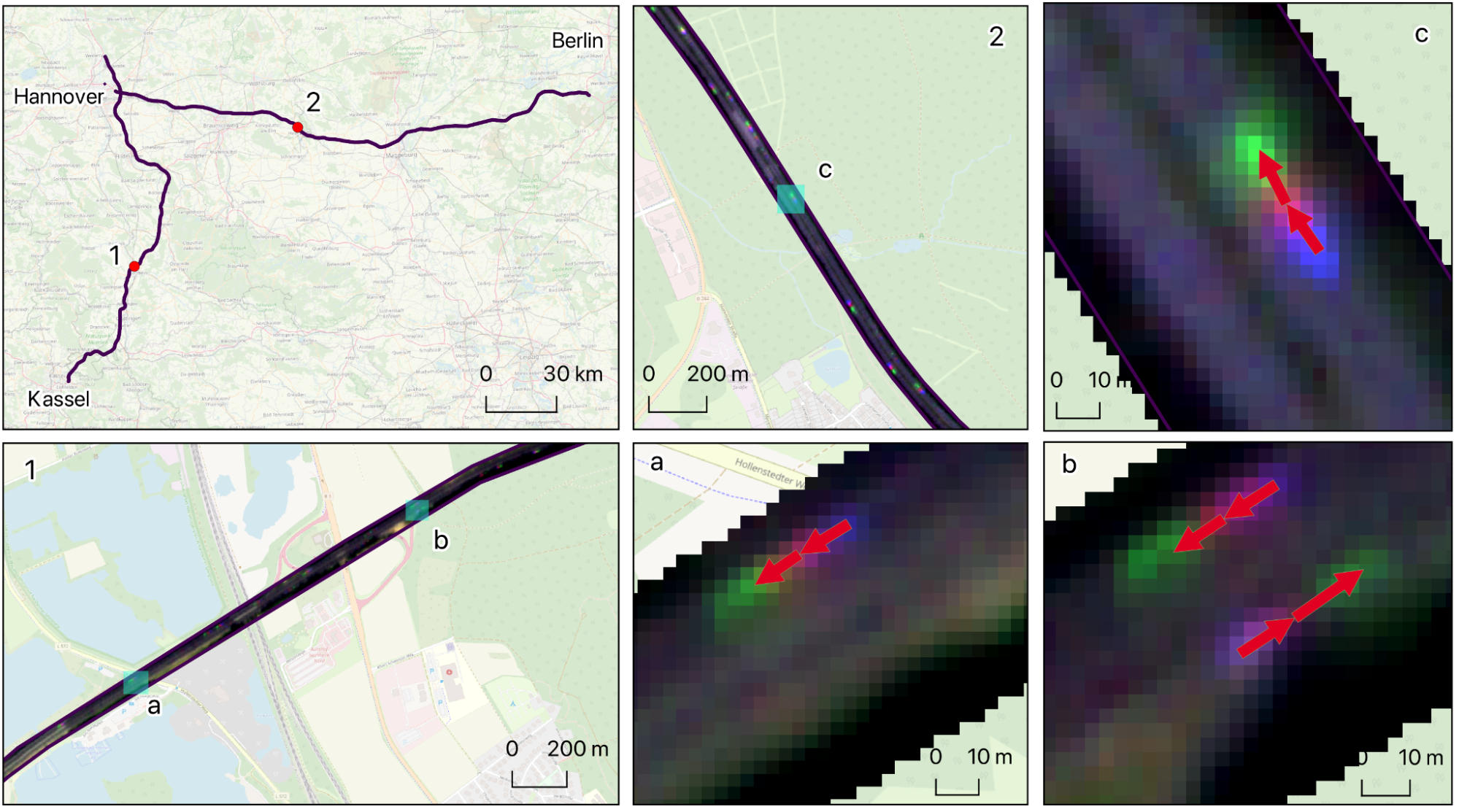}
    \caption{Annotation of moving vehicles on Highway A7 and A2 (Germany) based on RGB images (a, b, c) of PlanetScope SuperDove scenes from 26.09.2023.}
    \label{fig:MovingEchoesWithTrajectoriesAndBaseMap}
\end{figure*}

We use a pose estimation technique to detect the position and orientation of moving vehicles in the three-band images. Pose estimation models are trained on keypoints, marking elements, and their connectivity to other keypoints. We trained a model to detect keypoints corresponding to the 2D coordinates of the peak intensity areas that represent vehicles. Each vehicle is marked by three keypoints connected in a sequence that mirrors the temporal arrangement of SuperDove bands: Blue → Red → Green (Fig. \ref{fig:MovingEchoesWithTrajectories}). After manually preparing a training dataset, a Keypoint R-CNN model was implemented and fitted.
For the training dataset (Fig. \ref{fig:MovingEchoesWithTrajectoriesAndBaseMap}), road imagery was specifically extracted using OSM street centrelines with a 30-metre buffer to ensure the focus is on roads. Within these extracted areas, we manually identified in QGIS software the echoes of moving vehicles, and placed three keypoints for each vehicle, one per image band, as described above. The dataset comprises 3,236 vehicle echoes collected into 726 clipped satellite images, each containing at least one echo. This dataset was then randomly split into three distinct subsets: training (502 images), validation (77 images), and test (147 images).
\subsubsection{Keypoint correction}
\begin{figure*}[t]
    \centering
    \includegraphics[width=1\linewidth]{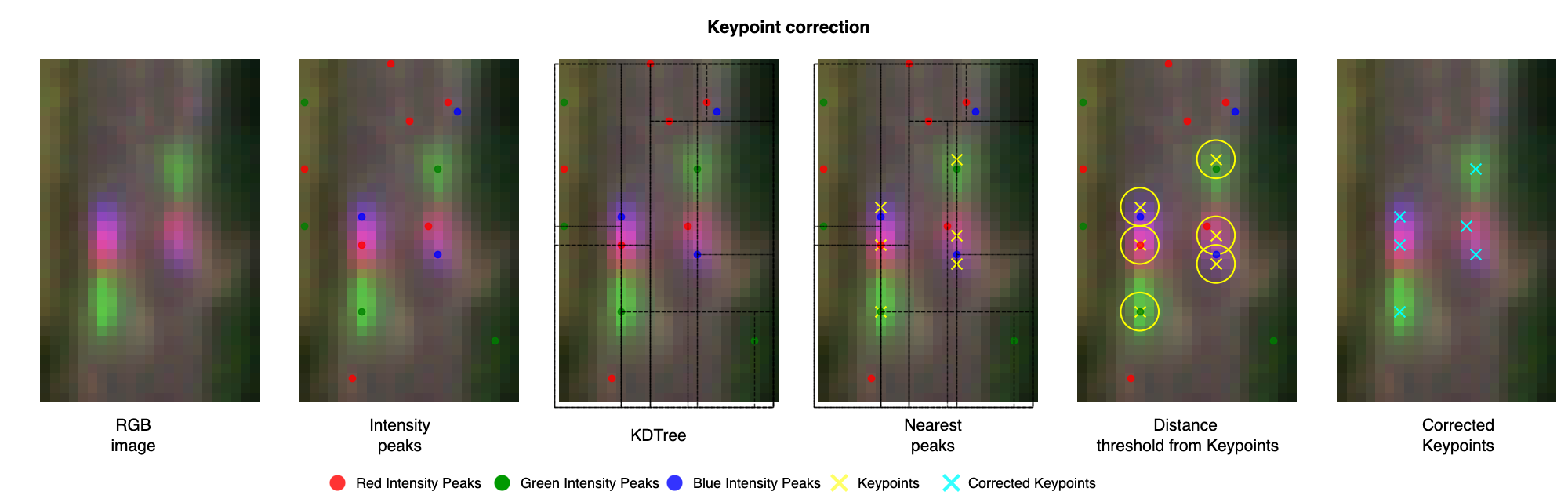}
    \caption{Keypoint correction algorithm visualization.}
    \label{fig:KeypointCorrectionAlgorithmVisualization}
\end{figure*}

We developed an algorithm to refine the positioning of keypoints by identifying the highest pixel intensities of individual vehicles and adjusting the keypoints to these peaks. This mechanism is utilized in two phases of our process. Initially, it enhances the annotation of vehicles by applying the algorithm before the keypoints are input into the neural network for training (Fig. \ref{fig:ProposedAlgorithm}). Subsequently, it increases the accuracy of the predicted keypoint locations by applying the algorithm again after the pose estimation model has been applied.
\\
The keypoint correction algorithm iterates over image bands to identify local intensity maxima, that is, peaks around pixels that are higher than their neighbouring pixels by at least h percent. We use a 3×3 cell neighbourhood, roughly equivalent to 11.1×11.1 metres in our context, and set h to 0.02 to identify significant maxima. The result is a binary image highlighting these maxima or intensity peaks per band. A k-d tree is then constructed based on the coordinates of these maxima. For each keypoint, the nearest coordinate in the k-d tree is determined and assigned as the new location of the keypoint, provided this location is within a threshold distance from its original position. We use a distance of two cells for this threshold, which translates to roughly 7.4 metres. If no suitable peak is found within this range, the original position of the keypoint is retained. Algorithm 1 details the step-by-step process of the keypoint correction mechanism (Fig. \ref{fig:KeypointCorrectionAlgorithmVisualization}).
\begin{algorithm}
\caption{Keypoint Correction}
\begin{algorithmic}[1]
\State \textbf{Input:} image\_channels, keypoints, max\_shift\_distance
\State \textbf{Output:} corrected\_keypoints

\State corrected\_keypoints $\gets$ empty list
\For{each channel $ch$ in image\_channels}
    \State maxima $\gets$ find intensity maxima in $ch$ using scikit's \texttt{h\_maxima}
    \State intensity\_peaks $\gets$ extract maxima coordinates
    \State kd\_tree $\gets$ build k-d tree from intensity\_peaks
    \For{each keypoint $kp$ in keypoints}
        \State nearest\_peak $\gets$ find nearest neighbour of $kp$ in kd\_tree
        \If{distance(nearest\_peak, $kp$) $<$ max\_shift\_distance}
            \State corrected\_kp $\gets$ nearest\_peak
        \Else
            \State corrected\_kp $\gets$ $kp$ \Comment{retain original keypoint}
        \EndIf
        \State add corrected\_kp to corrected\_keypoints
    \EndFor
\EndFor
\end{algorithmic}
\end{algorithm}
\subsubsection{Keypoint R-CNN pose estimation model}
We utilized a Keypoint R-CNN model to perform the keypoint estimation task (PyTorch, 2024). The neural network is based on the well-known Mask R-CNN \citep{he2018mask} model, incorporating a Feature Pyramid Network (FPN) and Residual Network (ResNet-50) backbone for extracting features. The weights of the model are initialized randomly and trained from scratch. The architecture is configured to detect two object classes (‘moving echoes’ and background) and predict three keypoints per moving vehicle. 
\\
The Region Proposal Network (RPN) uses a custom anchor generator to generate region proposals, and the entire backbone is optimized during training. We employ the Adam optimizer \citep{kingma2014adam} for efficient gradient-based optimization and use a ReduceLROnPlateau scheduler to dynamically adjust the learning rate based on validation loss performance. The model is optimized against a composite loss function that includes several components: loss for classification measures the accuracy of class label predictions; bounding box regression assesses how closely predicted bounding boxes match the ground truth; keypoint placement evaluates the precision of predicted keypoint positions; objectness score determines whether a proposed region contains an object of interest; and RPN box regression optimizes the coordinates of anchor boxes proposed by the RPN to improve detection. These components are summed to form a total loss.
\\
Although Keypoint R-CNN needs more time to train than its main alternative CenterNet \citep{zhou2019objects}, its implementation is curated in the official PyTorch \citep{pytorch2024pytorch} repository, making the architecture accessible and ensuring reproducibility.
\subsubsection{Model validation metrics}
\label{subsubsec:ModelValidationMetrics}
To assess the accuracy of our vehicle keypoint detection model, we compute two standard evaluation metrics used in the pose estimation. The first one is the Root Mean Square Error (RMSE), which is calculated based on the predicted and annotated trajectory lengths, and lower values indicate better model performance.
\\
The second metric, Mean Average Precision (mAP), is determined by averaging the Average Precision (AP) over all keypoints at Object Keypoint Similarity (OKS) thresholds ranging from 0.5 to 0.95 in 0.05 increments, where AP is derived from the area under the precision-recall curve (Equation \ref{eq:03MeanAveragePrecision}).
\begin{equation}
\text{mAP} = \text{mean}\left\{\text{AP@OKS}(0.50\!:\!0.05\!:\!0.95)\right\}
\label{eq:03MeanAveragePrecision}
\end{equation}
Generally, OKS is calculated from the distance between the predicted point and its ground truth, normalized by the scale of the object. Higher OKS indicates greater spatial overlap between predictions and ground truth keypoints.  Here, we employ a customized version of the OKS metric that accounts for the characteristics of fast-moving vehicles in our dataset. The standard OKS metric is computed as follows:
\begin{equation}
\text{OKS} = \frac{1}{N} \sum_{i=1}^{N} exp \left ({-\frac{d_i^2}{2s^2k_i^2}}\right )
\label{eq:03ObjectKeypointSimilarity}
\end{equation}
where $N$ is the number of keypoints, $d_i$ represents the Euclidean distance between the predicted and ground truth keypoints, $s_i$ is the scale of the object, set uniformly to 1 in our model for unbiased evaluation, and $k_i$ is a per-keypoint constant, traditionally a function of the object’s size. In our adaptation, we redefine $k_i$ to reflect the overall pixel distance among three connected keypoints that form a connected segment representing a single echo of a vehicle. This modification allows $k_i$ to account for the increased distances caused by vehicle speed, effectively normalizing these gaps. Additionally, while the OKS formula can incorporate a visibility parameter to account for the detectability of keypoints, our analysis does not include this variable, as all keypoints are unobstructed and clearly visible.
\\
The best-performing model was selected for its balanced performance, achieving a high mAP while also maintaining a low trajectory length RMSE. Additionally, only keypoints with scores above 0.7 were considered true positives in the mAP metric calculation, ensuring that our evaluation focuses on high-confidence detections.

%% file: chapters/03MaterialsAndMethods/03_04VelocityEstimation.tex
\subsection{Velocity estimation}
\label{subsec:04VelocityEstimation}
The process for determining the vehicle velocity begins by identifying three keypoints in real-world metre coordinates corresponding to the vehicle's position in each band's image, as described above. The mean distance travelled by the vehicle $d_{mean}$ is calculated by averaging the distances between the keypoints pairs for the blue and red bands, and the red and green bands. 
The next step (Equation \ref{eq:04ElapsedCaptureTimeBetwenBands} is to ascertain the time elapsed between the captures of these bands $\Delta t$:
\begin{equation}
\Delta t=\left ( \frac{v_{satellite}}{w_{bands}d_{GSD}} \right )^{-1}
\label{eq:04ElapsedCaptureTimeBetwenBands}
\end{equation}
where $v_{satellite}$ is the satellite's velocity (m/s), $w_{bands}$ represents the number of pixels across the band's width, and $d_{GSD}$ is the instrument ground sampling distance (m/px). We assume that $v_{satellite}$ is constant across the recording of the bands. $w_{bands}$ is identical for the different bands \citep{planet2024planetscope}. For $d_{GSD}$, we consider the average values between the bands. Thus $\Delta t$ is representative of each individual SuperDove band. 
The final formula (Equation \ref{eq:04VehicleVelocityCalculation}) for calculating the vehicle's velocity $V$ is then a simple ratio of the distance the vehicle has covered to the time interval over which that distance was covered:
\begin{equation}
v= \frac{d_{mean}}{\Delta t}
\label{eq:04VehicleVelocityCalculation}
\end{equation}
This procedure assumes the following: First, the bands’ peak distance is independent of the direction of motion. Second, the Satellite velocity acquired by querying Planet ephemeris state vector \citep{planet2024planetscope} is constant across a specific selected date. Third, initial object acceleration is assumed to be zero. In the following sections, we describe the validation of our approach.

%% file: chapters/03MaterialsAndMethods/03_05VelocityValidation.tex
\subsection{Velocity validation}
\label{subsec:05VelocityValidation}
The assessment of the accuracy of speed estimation is challenging due to the absence of ground truth information on vehicle speed. In this study, we propose a novel method to validate the accuracy of estimated vehicle velocities derived from object echoes, utilising two distinct estimates, based on UAV video and GPS trajectories.
\subsubsection{Velocity estimation from UAV videos}
\label{subsubsec:05VelocityEstimationFromUavVideos}
The first experiment employed drone video footage from a DJI M3T. Velocity estimation in UAV videos involves detecting a vehicle's position within a given frame, tracking the vehicle across frames, and calculating their velocities by measuring how far they move within each frame. For vehicle detection, we utilized a custom YOLOv8 \citep{jocher2023ultralytics} object-detection model designed to identify objects of predefined categories (e.g., cars and pedestrians) from images taken by drones \citep{shamrai2024yolov8l}. The model was trained on the COCO dataset \citep{lin2014microsoft} and fine-tuned with the VisDrone dataset, captured by various drone-mounted cameras \citep{zhu2021detection}. The model has a reported vehicle detection mAP50 score of 0.46137 \citep{shamrai2024yolov8l}.
In the subsequent phase, we implemented the ByteTrack algorithm \citep{zhang2021bytetrack} to track vehicles across video frames. This algorithm utilizes a data association technique, which links detection boxes across subsequent frames by analyzing their similarities. In our application, boxes representing vehicles detected by the customized YOLOv8 model are associated across subsequent frames of drone-captured footage.
Detecting and tracking vehicles through video analysis enables the calculation of their velocities by measuring how far they move within each frame. To scale these pixel-based measurements into a real-world metric, we first derive the Ground Sampling Distance (GSD) for our image (Equation \ref{eq:03GroundSamplingDistance}). The GSD depends on the focal length (F) and dimensions of the sensor (S) of the camera, as well as the altitude (A) at which the drone is operating, and the dimensions of the image (I). It is calculated for both the width (w) and height (h) directions of the image as follows:
\begin{equation}
 GSD(w,h)=\frac{A\times S(w,h)}{F\times I(w,h)}
\label{eq:03GroundSamplingDistance}
\end{equation}
The specific values utilized in our experiment are provided as follows: the focal length $F$ of the drone camera is 4.4 mm, and the sensor dimensions $S$ are 6.4 mm in width $w$ and 4.8 mm in height $h$. These camera specifications are documented in the DJI Mavic 3T's technical sheet. The image dimensions remain constant at a resolution of, 8000×6000 pixels. Altitude data is dynamically captured in the video footage by the drone's onboard GPS.
Using the $GSD$, we calculate the travelled distance $D_{i}$ of the $i$-th vehicle between its initial $f_{m}$ and final $f_ {n}$ frames in the image:

\begin{align}
 D_{i}\left ( f_{n},f_{m} \right ) = & \sqrt{\left [ \left ( C_{x,f_{n}}-C_{x,f_{m}} \right )\times GSD_{w} \right ]^{{^{2}}}} \notag \\
 & + \sqrt{\left [ \left ( C_{y,f_{n}}-C_{y,f_{m}} \right ) \times GSD_{h} \right ]^{{^{2}}}}
\label{eq:03TravelledDistanceCarOnFrames}
\end{align}

where $C_{x,f_{m}}$ and $C_{y,f_{m}}$ represent the $x$ and $y$ coordinates of the $i$-th vehicle’s centroid at the initial frame $f_{m}$, and $C_{x,f_{n}}$ and $C_{y,f_{n}}$ represent the coordinates in the final frame $f_{n}$.  $GSD_{w}$ and $GSD_{h}$ denote the ground sampling distances on the $x$ and $y$ axes, respectively. The initial and final frames are used, assuming that vehicle speeds on highways are consistent over short distances and thus representative of the vehicle's velocity along the 620m highway segment.
Finally, the velocity $v_{i}$ of the vehicle is calculated (Equation \ref{eq:03VehicleVelocity}) by dividing $D_{i}$ by the time interval  $N_{i} \left (f_{n} , f_{m} \right )$, which is the number of frames between $f_{m}$ and $f_{n}$, adjusted by the frame rate $\left (fps \right )$:
\begin{equation}
v_{i}=\frac{D_{i}}{N_{i}\left (f_{n},f_{m} \right )}\times fps
\label{eq:03VehicleVelocity}
\end{equation}
After deriving the velocities of vehicles from the videos, we compared them to the velocities estimated from SuperDove imagery. For this purpose, one hundred SuperDove images (March 2023 – March 2024) representing the target highway segment were manually annotated, and their velocities calculated as described above. The satellite dataset includes velocity estimations for 307 moving vehicles. 
\\
To compare velocity measurements from drone footage and SuperDove satellite imagery, we analyzed their distributions using boxplots and histograms, along with descriptive statistics and a Kolmogorov-Smirnov (KS) test. We report the KS statistic, which is the maximum difference between the cumulative distribution functions (CDFs) of the two datasets, as determined by the KS test. We also report its p-value, which indicates the evidence against the null hypothesis that the two datasets are from the same distribution.

\subsubsection{Velocity estimations from GPS traffic data}
\label{subsubsec:VelocityEstimationsFromGpsTrafficData}
The second experiment involves the comparison of the SupverDove speed estimates with speed values derived from GPS car trajectories. We visually searched SuperDove images for GPS trajectories (from 15 March 2020 to 30 April 2024), using QGIS software. We selected only trajectories that fall within the date range for which we also have images.
\begin{figure}[H]
    \centering
    \includegraphics[width=0.8\linewidth]{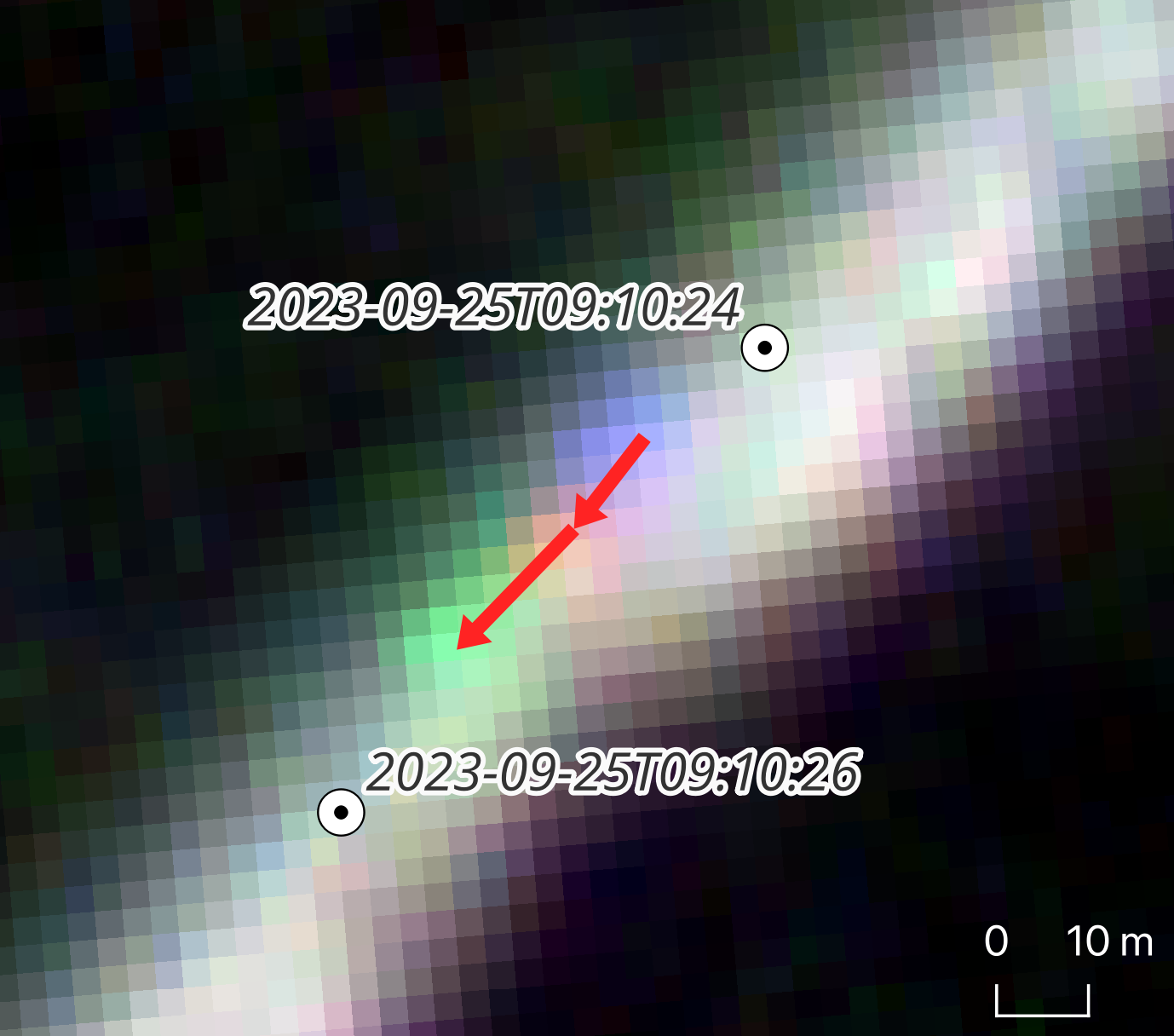}
    \caption{EnviroCar planet data matching based on GPS trajectories.}
    \label{fig:03EnvirocarPlanetDataMatchingBasedOnGpsTrajectories}
\end{figure}
The acquisition of GPS tracks and their matching echoes in SuperDove images involved applying several filters to both the trajectory data and the corresponding Planet scenes. Filters were applied to the GPS tracks to retain only the morning hour records. Since Planet images are captured between 9:00 am and 11:00 am, GPS records outside this time frame were excluded to ensure temporal alignment with the Planet imagery. A 10-metre buffer was then created around each trajectory to define the area of interest for acquiring the corresponding Planet images. For the Planet scenes, filters were applied to select high-quality images, considering factors such as cloud cover (up to 50 \%) and visibility of the area of interest (up to 50 \%). These measures ensured that only the most relevant tracks and images were used for subsequent analysis of moving echoes. 
\\
GPS points of individual trajectories, labelled with the timestamp of their recording, were overlaid with SuperDove RGB composites of the same dates. We matched each GPS point with the corresponding vehicle echo in the SuperDove images, confirming that both the spatial positions and the timestamps aligned closely (Figure \ref{fig:03EnvirocarPlanetDataMatchingBasedOnGpsTrajectories}).
\\
We then applied the SuperDove-based velocity estimation method to calculate the speeds of the vehicles identified in both the GPS data and the SuperDove images, through annotation, and directly compared these measurements. The resulting difference between this estimation for 13 vehicles and their GPS velocity of the trajectory serves as the comparison indicator.
\\

%% file: chapters/04Results.tex
\section{Results}
\label{sec:results}
\input{chapters/04Results/04_01ExperimentalSettingsAndImplementationDetails}
\input{chapters/04Results/04_02VelocityEstimationEvaluation}

%% file: chapters/04Results/04_01ExperimentalSettingsAndImplementationDetails.tex
\subsection{Keypoint detection}
\label{04:KeypointDetection}
\subsubsection{Experimental settings and implementation details}
\label{04:ExperimentalSettingsAndImplementationDetails}
The computational experiments were conducted using NVIDIA 3090Ti, 1080, and A4000 GPU accelerators. The PyTorch Lightning library, an advanced wrapper for PyTorch that simplifies the management of complex neural network training processes, was utilized to ensure consistency and scalability of the training procedure \citep{pytorch2024pytorch}. During the process, \citet{neptune2019neptune} software was used to track the experiments. 
\\
\begin{figure*}[t]
    \centering
    \includegraphics[width=0.9\textwidth]{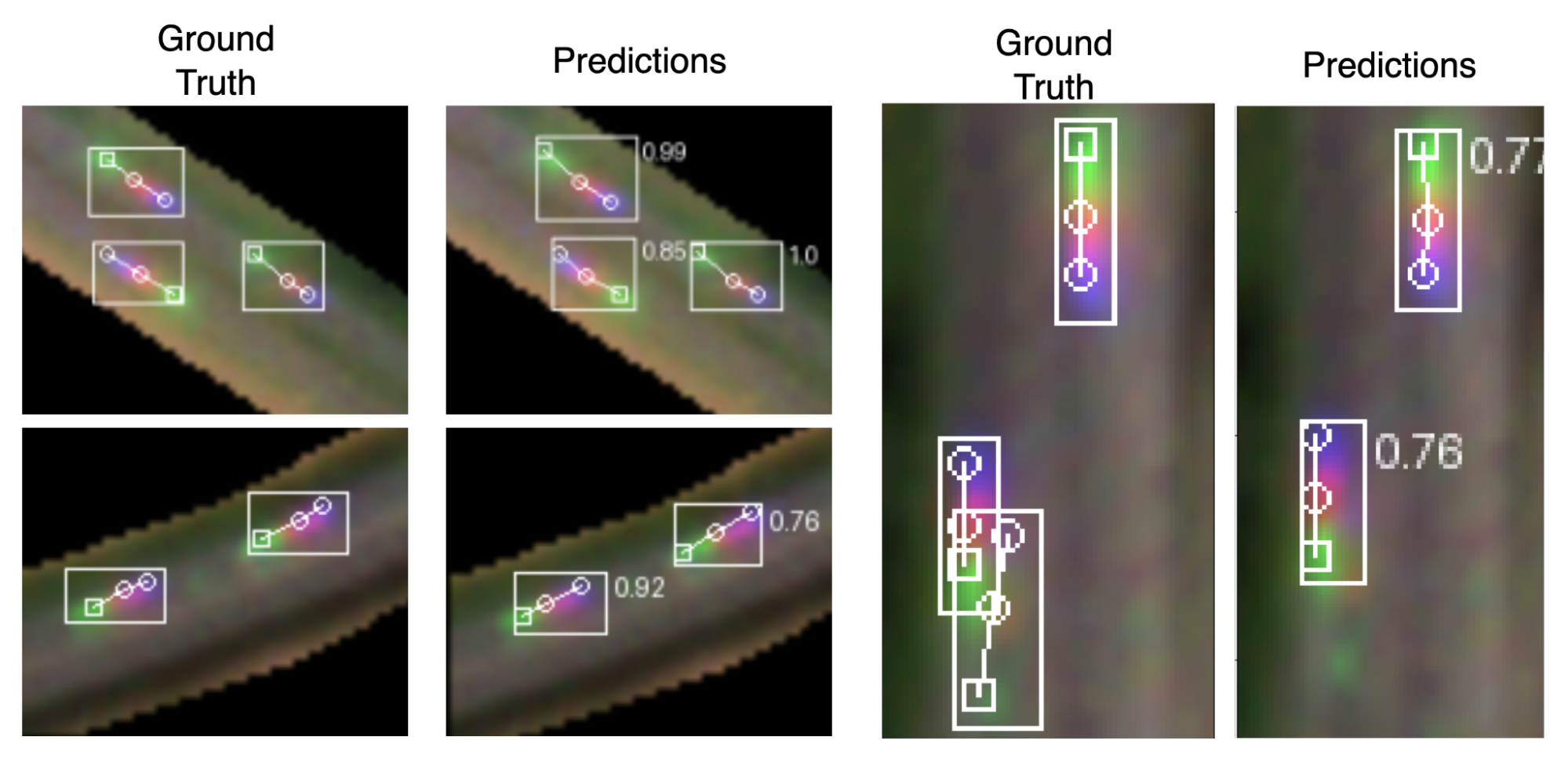}
    \caption{Model evaluation results: ground truth with key points and  bounding boxes, and predictions with confidence level.}
    \label{fig:04ModelEvaluationResultsGroundTruthKeypoints}
\end{figure*}
Before the main evaluation, approximately 120 experimental runs were scheduled to fine-tune various hyperparameters. For the final model training, five rounds of training were adopted, each utilizing randomized data sampling to mitigate biases that might arise from the order or composition of the training data. A scheduled learning rate was set up, starting at 0.001 and decaying to 0.00001 to accommodate the diminishing returns on learning as the model approached optimal solutions. Furthermore, gradient clipping was employed with a threshold set to 1.5, to enhance the stability of training in scenarios prone to erratic gradient values.
\\
After applying the keypoint correction algorithm to the training dataset, it was augmented by applying random rotation, vertical and horizontal flipping, brightness correction and perspective transformation. 
\subsubsection{Keypoint detection results}
\label{subsubsec:KeypointDetectionResults}
The dataset is divided into three groups: training, validation and test dataset, by 80\%, 10\% and 10\% correspondingly. The image size is 512 pixels. While training the model, we apply k-fold cross validation to select the best model, where $k = 5$.  In this study.
\\
An individual Keypoint R-CNN model training on PlanetScope satellite imagery run took 2 hours and 30 minutes on average to complete. Evaluation results are presented in table 1. The model achieved, at best, a 0.5925 mAP for the test subset. This is reasonable, considering the application of a restrictive prediction score filtering.
\begin{table}[H] 
\caption{Keypoint R-CNN model evaluation results trained on five randomly sampled dataset folds.}
\label{tab:04KeypointRCNNModelEvaluationResults}
\centering
\resizebox{\columnwidth}{!}{%
\begin{tabular}{|c|c|c|c|c|c|c|}
\hline
Fold & $loss_{val}$ & $loss_{test}$ & $mAP_{val}$ & $mAP_{test}$         & $RMSE_{val}$ & $RMSE_{test}$        \\ \hline
1    & 4.8534       & 4.8423        & 0.6017      & \textbf{0.5925}      & 5.2171       & 2.5301               \\ \hline
2    & 5.0503       & 5.189         & 0.4517      & 0.4228               & 2.9589       & 2.8983               \\ \hline
3    & 5.1667       & 5.1719        & 0.4426      & 0.3949               & 3.1157       & 2.3554               \\ \hline
4    & 5.2576       & 5.297         & 0.5875      & 0.5841               & 1.7889       & \textbf{1.9063}      \\ \hline
5    & 5.0124       & 4.8021        & 0.4041      & 0.4079               & 2.5987       & 3.389                \\ \hline
\end{tabular}%
}
\end{table}
Regarding travelled distance calculation, the model exhibits a minimal root mean squared error (RMSE) of 1.9063 px. Given 3.2–4.0 m GSD, the lowest velocity estimation error is on average around 7.4 m/s (26.28 km/h). Visualization of the best model results is presented in Figure \ref{fig:04ModelEvaluationResultsGroundTruthKeypoints}. Predictions of moving vehicle with confidence level greater than \textgreater 0.7 are presented.  
\\

%% file: chapters/04Results/04_02VelocityEstimationEvaluation.tex
\subsection{Velocity estimation evaluation}
\label{subsec:04VelocityEstimationEvaluation}
\subsubsection{Drone data comparison}
\label{subsubsec:04DroneDataComparison}
\begin{figure*}[t]
    \centering
    \includegraphics[width=1\linewidth]{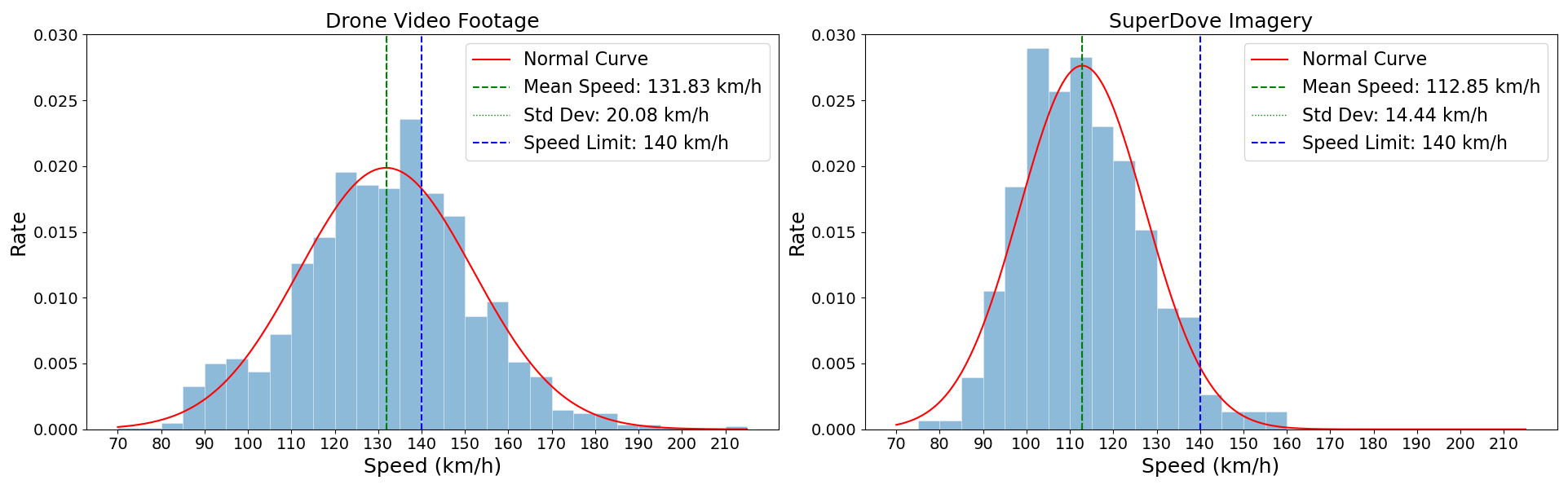}
    \caption{Speed distributions based on drone footage and SuperDove imagery.}
    \label{fig:04SpeedDistributions}
\end{figure*}

Figure \ref{fig:04SpeedDistributions} presents the speed distribution curves from drone footage and satellite imagery. Both datasets exhibit characteristics of normal distribution patterns, as confirmed visually and supported by their respective kurtosis and skewness values. The drone data shows a kurtosis of 0.292 and a skewness of 0.070, indicating a very slight peaking and symmetric distribution around the mean. The satellite-based data has a kurtosis of 1.280 and a skewness of 0.102, suggesting a more pronounced peak and a similarly slight rightward skew, though still maintaining a generally symmetric distribution. The Shapiro-Wilk test results further support these observations, with W=0.9956 and p=0.00013 for the drone data, and W=0.9807 and p=0.00040 for the satellite-based data, indicating slight deviations from normality in both cases.
\\
However, the curves reveal distinct differences in their statistical profiles. The drone data, with a mean of 131.83 km/h, is centred just below the speed limit of 140 km/h, aligning well with expectations for nearly free-flowing highway traffic. The standard deviation is 20.08 km/h, indicating a broad spread of speeds ranging from a minimum of 71 km/h to a maximum of 214 km/h. 
\\
In contrast, the satellite-based data is centred around a lower mean of 112.85 km/h, with a narrower standard deviation of 14.44 km/h. The speeds in this dataset range from 45 km/h to 158 km/h. The narrower range and lower mean suggest that the satellite method might be compressing the variability in speed measurements, potentially leading to underestimations, particularly at higher speeds. 
\\
A Kolmogorov–Smirnov test confirms these observed differences, yielding a statistic of 0.46 and a p-value of 4.85e-50, indicating a significant difference between the two datasets and supporting the conclusion that they stem from distinct distributions.
\subsubsection{GPS data feed comparison}
\label{subsubsec:04GprsDataFeedComparison}
\begin{figure}
    \centering
    \includegraphics[width=1\linewidth]{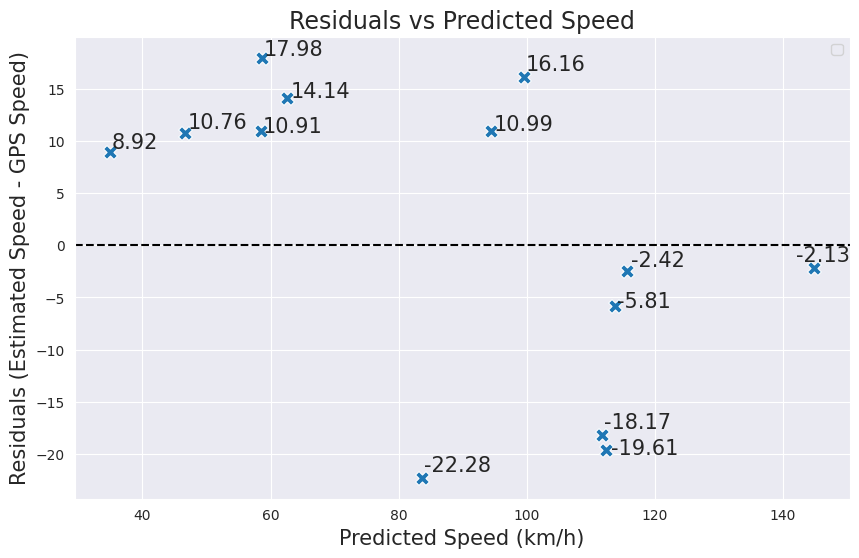}
    \caption{Residual plot showing the predicted velocity of the 13 vehicles in relation to the prediction residuals.}
    \label{fig:04ResidualsPlot}
\end{figure}

Figure \ref{fig:04ResidualsPlot} presents a scatter plot of the predicted velocity of the 13 vehicles versus the prediction residuals, which are defined as the difference between the estimated and GPS-measured velocities.
\\
The residuals tend to decrease as the predicted speed increases. This trend suggests the presence of a systematic bias in our model. At lower predicted speeds, the model tends to overestimate speeds (positive residuals), while at higher predicted speeds, the model tends to underestimate speeds (negative residuals).  For instance, at the anticipated velocities between 40 and 100 km/h, the residuals are predominantly positive, with one outlier exhibiting a markedly negative value. This indicates an overestimation, with residuals ranging from approximately -22.86 km/h to 17.98 km/h, with a standard deviation of 13.81 km/h. Conversely, at a predicted velocity from 100 km/h and above, residuals ranging from -2.13 to -19.61 km/h and a standard deviation of 8.59 km/h indicate a significant underestimation by the model.

%% file: chapters/05DiscussionAndConclusion.tex
\section{Discussion and Conclusion}
\label{sec:discussionandconclusion}
Vehicle speed estimation is essential in traffic management operations to monitor traffic flow and enforce speed limits. Our method is automatic and applicable to any medium-resolution satellite imagery. Therefore, it is scalable and can be applied globally with daily coverage, unlike conventional traffic monitoring methods and manual satellite-based approaches. The validation framework supports adaptation of the method to different environments and conditions. The method is thus useful for modelling transportation systems, analysis of traffic patterns, monitoring of congestion levels, road safety and sustainable transport systems.
\\
The proposed model for extracting moving vehicles demonstrated reasonable performance, achieving a mAP score of approximately 0.53 at a confidence level of 0.7. Although this score is lower than many published models, it is important to consider that we are working with vehicles detected in medium-resolution satellite imagery. As expected, detection accuracy for smaller objects is considerably lower than for larger ones \citep{tong2022deep}, but the results remain sufficient for further velocity estimation analysis.
\\
For travel distance, the proposed model exhibits an RMSE of up to 1.9063 pixels, corresponding to approximately 3.3 metres at ground level, which translates into a velocity of 7.3 m/s. This level of precision demonstrates the model's capability to provide reasonably accurate estimates, despite the inherent limitations of satellite-based data. Furthermore, the minimum detected ground speed, derived from manual labelling of motion anomalies, is approximately 11 m/s, which is notably more precise than the 20 m/s reported by \citet{keto2023detection}.
\\
The speed validation process was conducted in two ways. The first, more conventional method involved comparing the predicted speeds with GPS tracks (Fig.10). When compared to the EnviroCar trajectories, the average vehicle speed estimation error is approximately 3.4 m/s. However, the model exhibits systematic bias, overestimating at lower predicted speeds below $\sim$28 m/s, and underestimating at higher ones. We only had 13 cars in this analysis due to the difficulty of finding corresponding vehicles between GPS tracks and satellite images, and a more comprehensive comparison is needed, which can be explored in future work.
\\
The validation based on drone footage (Fig.\ref{fig:04SpeedDistributions}) exhibits a consistently normal distribution. However, the consistent underestimation of speed indicates that calibration is necessary when using PlanetScope data for precise speed measurement applications. One explanation might be attributed to the differences in the spatial and temporal resolutions of these two sensing technologies. In comparison to the satellite, drones can continuously capture detailed, high-resolution data that explain minor variations in speed more accurately. In contrast, the coarser spatial and temporal resolution of satellite imagery might smooth over these variations, leading to underestimate of speeds. Another explanation could be that we systematically underestimate velocity due to how we calculate the travelled distance. We proved experimentally that consistently increasing the distance by one GSD (3–4 m) for each sample results in similar distributions. Other potential reasons for errors in speed estimation include weather conditions, light exposure, clouds, road type, and the vehicle's colour, size, and speed.
\\
Integrating our method with existing traffic flow data sources, can enhance traffic feed data by providing a spatially continuous, and large-scale view of traffic volume and speed. This would be particularly valuable in regions where conventional traffic data is sparse or unavailable. For instance, GPS-based car probes typically have penetration rates of up to 5\% of the total traffic flow. Our method can be used to fill in the gaps, and fusing these sources merits dedicated research. Applications include large-scale estimation of traffic density and congestion. In addition, effective navigation systems, such as openrouteservice \citep{openrouteservice2024} require reliable localized speed limit estimates, and these are not always publicly available, especially in less developed countries. Our method might be extended to detection of slow and clustered vehicles as indicators of traffic congestion. Additionally, future work will focus on making derived data and applications available on a broader scale with licences that allow for non-commercial use and derived work.
\\
Application of our method daily, and at a global scale, requires further work. The variability of road networks worldwide may affect performance, as demonstrated by \citet{fisser2022detecting}. Another obstacle is the extraction of daily satellite imagery for the global road networks ($\sim$65,000,000 km), which requires efficient methods. With a 30-metre buffer, the estimated coverage area for road networks is $\sim$2,000,000 km$^{2}$ for the world. PlanetScope images, with their high-resolution data, allow for the detection of vehicles even across these large areas.
\\
Through large-scale, regional and global observation capabilities, PlanetScope's mission empowers researchers to monitor traffic dynamics across vast geographical expanses, including remote and inaccessible regions, providing crucial insights into temporal variations and trends essential for understanding dynamic transportation systems. With high spatial resolution, PlanetScope imaging facilitates precise tracking of individual vehicles, enabling microscale analysis of traffic flow, traffic behaviour and localized phenomena such as congestion or accidents. Moreover, the consistency and longitudinal nature of satellite-based data support in-depth studies of traffic trends, changes, and impacts in transportation network systems. Our insights of the proposed novel approach to estimate moving vehicle velocity on highways is beneficial to transportation systems, particularly in areas with low coverage of conventional traffic data and low presence of technologies. We intend to apply the method on the global scale to extract daily information on vehicles’ velocity on highways.

%% file: chapters/06AuthorContributions.tex
\section*{Author Contributions}
\label{sec:authorcontributions}
Conceptualization, Adamiak, M., Grinblat, Y., and Psotta, J.; methodology,  Adamiak, M.; coding,  Adamiak, M., and Psotta, J.; data processing, Adamiak, M., Grinblat, Y., and Psotta, J.; analysis, Adamiak, M., Psotta, J., Tang, S., and Mazumdar, H.; writing, Adamiak, M., Grinblat, Y., Psotta, J., Fulman, N.; supervision, Zipf, A.; All authors have read and agreed to the published version of the manuscript.

%% file: chapters/07Acknowledgement.tex
\section*{Acknowledgements}
\label{sec:acknowledgements}
The authors extend their gratitude to the Klaus Tschira Stiftung (KTS) for core funding of HeiGIT.
Nir Fulman was supported by the Health + Life Science Alliance Heidelberg Mannheim and received state funds approved by the State Parliament of Baden-Württemberg.

%% file: chapters/08DeclarationOfCompetingInterest.tex
\section*{Conflicts of Interest}
\label{sec:conflictsofinterest}
The authors declare no conflict of interest.

%% file: chapters/09References.tex
\section*{References}
\flushend
\renewcommand{\refname}{}
\vspace{-6mm}

\bibliographystyle{agsm}
\bibliography{bibliography.bib}